\documentclass{article} % For LaTeX2e
\usepackage{iclr2025_conference,times}
\usepackage{amsthm, amsmath}
\usepackage{xcolor}
\usepackage{algpseudocode, algorithm}
\usepackage{graphicx}
\usepackage{wrapfig}
\usepackage{multirow}
\usepackage[caption=false]{subfig}
\usepackage[export]{adjustbox}

\theoremstyle{definition}
\newtheorem{definition}{Definition}[section]

\theoremstyle{remark}

\usepackage{amsmath,amsfonts,bm}

\def\eqref#1{equation~\ref{#1}}
\def\1{\bm{1}}

\DeclareMathAlphabet{\mathsfit}{\encodingdefault}{\sfdefault}{m}{sl}
\SetMathAlphabet{\mathsfit}{bold}{\encodingdefault}{\sfdefault}{bx}{n}

\newcommand{\mc}[1]{\mathcal{#1}}
\newcommand{\mb}[1]{\mathbb{#1}}

\usepackage{hyperref}
\usepackage{url}

\newcommand{\model}[0]{m_\theta}
\newcommand{\PAtrain}[0]{\text{PA}_{\text{train}}}
\newcommand{\PAss}[0]{\text{PA}_{\text{ss}}}

\title{Data Efficient Subset Training with \\Differential Privacy}

\author{Ninad Jayesh Gandhi$^*$, Moparthy Venkata Subrahmanya Sri Harsha\thanks{equal contribution} \\ 
Centre for Machine Intelligence and Data Science\\
IIT Bombay\\
\texttt{\{ninad.gandhi,24M2151\}@iitb.ac.in} \\
}

\iclrfinalcopy % Uncomment for camera-ready version, but NOT for submission.
\begin{document}

\maketitle

\begin{abstract}
    Private machine learning introduces a trade-off between the privacy budget and training performance. Training convergence is substantially slower and extensive hyper parameter tuning is necessary. Consequently, efficient methods to conduct private training of models have been thoroughly investigated in the literature. To this end, we investigate the strength of the data efficient model training methods in the private training setting. We adapt GLISTER \citep{glister} to the private setting and extensively assess its performance. We empirically find that practical choices of privacy budgets are too restrictive for data efficient training to work in the private setting. We make our code publicly available \href{https://github.com/harsha-moparthy/DP-SubSel/tree/main}{here}.
\end{abstract}

\section{Introduction}
\label{sec:introduction}
Machine learning models often memorize training data \citep{extracting-training-data-1, extracting-training-data-2}. In many applications, such as healthcare, finance and generative AI, ensuring privacy of the dataset participants is of utmost importance. Historically, many heuristic methods have been attempted at providing privacy to the dataset participants such as anonymization of the data or removing sensitive columns. These methods have been shown to fail spectacularly in presence of an adversary that can perform \textit{linkage attack} \citep{dwork2014algorithmic} using auxiliary data and reconstruct significant portions of the dataset \citep{reconst-attacks}. A systematic study in the field of private machine learning was enabled by \textit{differential privacy} due to \citet{dp-first-defn}.

\begin{definition}[$(\varepsilon, \delta)$- Differential Privacy]
    \label{defn:DP}
    A randomized mechanism $\mc{M}: \mc{D} \to T$ is $(\varepsilon, \delta)$- differentially private, if $\forall x,x' \in \mc{D}$, such that $|x-x'|_1 \le 1$ and $\forall S \subseteq T$, we have that
    
    \begin{equation*}
    \mb{P} \left[ \mc{M}(x) \in S \right] \le e^\varepsilon \left[ \mc{M}(x') \in S \right] + \delta
    \end{equation*}

\end{definition}

Where, $|x - x'|_1$ is the $l_1$-norm of the datasets $x, x'$ and the unity bound indicates that they differ in at most one record. $\varepsilon$ and $\delta$ are the privacy loss parameters, higher value indicating lower privacy. By definition, differential privacy ensures that the presence or absence of a single entry in the dataset does not affect output of the mechanism \textit{significantly}. The private analysis in case of machine learning is the computation of gradient with respect to the model weights per sample.

Differential privacy has found large scale adoption in deep learning after the development of the DP-SGD algorithm \citep{dp-sgd}. DP-SGD uses \textbf{gradient clipping} and \textbf{noising} to induce privacy in training process and a \textit{privacy accountant} tracks the degradation of privacy throughout the training run. With DP-SGD, a model can be trained to achieve decent performance with modest privacy parameters $\varepsilon = 3$ and $\delta \le 1/|D_{train}|$. Though, DP-SGD algorithm poses a significant challenge due to sample gradient clipping which obliterates parallelism by effectively making the batch size equal to 1. Also, large scale problems such as ImageNet classification remain challenging in the private setting \citep{dp-randp}.

In the non-private setting, \textit{data efficient} model training has found much success. It has been shown to maintain the model performance while requiring less data to train. In light of this, we explore the data efficient training paradigm in the private setting. We thoroughly test this paradigm and report our empirical findings here:
\begin{itemize}
    \item The operations required to extract a high quality training data subset release private information and their privacy budget must be accounted for. Practical privacy budgets $\varepsilon\in [3,8]$ probe to be extremely restrictive and render the methods for data efficient training impractical.  
    \item We empirically show that the choices of privacy budgets make the search for quality data inefficient and also discuss conditions under which such methods can work. 
\end{itemize}

% The line of work termed as data efficient learning targets model training with lower data costs. For non private machine learning model training it has been shown that it suffices to train on a representative subset of the train data when there is redundancy among examples without losing much on performance but converging much faster \cite{TBD}. Our work explores the union of these two fields, that is data efficient machine learning with differential privacy. In our experiments, we find that this approach though interesting does not show good results. We provide some explanations for the same and empirically show the validity of our arguments.

% \begin{figure}[H]
%     \centering
%     \includegraphics[width=1.0\linewidth]{slide.png}
%     \caption{Flowchart of GLISTER-DP}
%     \label{fig:enter-label}
% \end{figure}

\section{Related Work}
\label{sec:related-work}

\textbf{Private Machine Learning.}
The composition theorems of differential privacy \citep{dp-composition-thms} provide components for building more complex mechanisms using simpler ones. The approach taken in most machine learning applications is that of privatizing gradients. DP-SGD \citep{dp-sgd} first provided a practical implementation of private machine learning, also designing a privacy degradation tracker termed as a \textit{privacy accountant} based on \textit{Rényi divergence} \citep{rdp}. The work by \citet{prv-accountant} develops a faster algorithm to approximate the bound for k-fold composition of homogeneous DP mechanisms in $O(\sqrt{k})$ time. \citet{dp-google-kurakin} show that private training performance depends on various factors; larger models are hard to train, hyperparameters tuning is essential and methods like transfer learning boost performance. \citet{de2022unlocking} show improvements in performance for training larger models. Moreover, large scale private training such as ImageNet classification remains a challenging task \citep{dp-randp} with SOTA test accuracy being just 39.39\% for $\varepsilon=8$. \citet{dp-tan-burn} introduce TAN, Total Amount of Noise during training, and use it to inform hyperparameter search for private training. \citet{dp-randp} achieve state of the art performance on multiple datasets across various choices of $\varepsilon$ by phased training with priors learned on noise generated by random processes.

% As discussed previously, private training differs from non private ML training in two aspects. 1. To reduce the statistical significance of gradients derived from any example, it's influence is bounded by \textbf{gradient clipping} and \textbf{multidimensional Gaussian noise} is added to it. 2. Given set privacy budget, the number of training steps is fixed. Noisy gradients slow down model convergence, and the privacy budget restricts the amount of training. There is a natural trade-off between privacy and utility of private training. Moreover, highly accurate private models for large scale tasks such as ImageNet classification remains a challenging task \cite{dp-randp}. DP training provides multiple challenges from model selection, to choice of hyperparameter. Smaller models provide better performance, large batch sizes are preferred and transfer learning provides accuracy boost \cite{dp-google-kurakin}. \cite{de2022unlocking} improve performance of larger models. \cite{dp-tan-burn} introduce TAN, Total Amount of Noise during training, and use it to inform hyperparameter search for private training. \cite{dp-randp} achieve state of the art performance on multiple datasets across various choices of $\varepsilon$ by phased training with priors learned on noise generated by random processes. 

\textbf{Data Efficient Training. }
Multiple approaches for data efficient model training have been investigated. One line of work explores iterative subset selection and training approaches and the goal is to find a high quality subset to train \citep{glister, CRAIG, CREST, GRADMATCH}. Searching for a high quality subset is a combinatorial problem which is generally solved by optimizing a submodular proxy function. This approach has been used in various domains of machine learning including speech \citep{data-eff-tr-submod-speech}, vision \citep{data-eff-tr-submod-vision} and natural language \citep{data-eff-tr-submod-nlp}.  Another line of work explores dataset distillation \citep{data-distillation, data-distillation-transformers}. Yet another line of methods exist exploring dataset pruning by retaining important examples based on their importance scores. Importance score of an example is a function of how often the example is forgotten throughout the course of training \citep{forget-scores, forget-scores-2}. Our work aligns with methods for searching a high quality subset to train models in the private setting.

% Data efficient machine learning is tackled through multiple approaches, such as identifying a representative data subset \citep{glister,CRAIG, CREST}, dataset distillation \cite{data-distillation, data-distillation-transformers}. Identification of a representative subset for training is tackled through a submodular proxy for the original objective. This approach has been used in various domains of machine learning including speech \cite{data-eff-tr-submod-speech}, vision \cite{data-eff-tr-submod-vision} and natural language \cite{data-eff-tr-submod-nlp}. Multiple works extract the training subset by gradient matching \cite{CRAIG, glister, gradmatch}. The subset selection and training happen iteratively in these methods and the subset adapts to the current state of the ML model. Other methods such as \cite{forget-scores, forget-scores-2} prune the dataset with important examples based on their importance scores. Importance score of an example is a function of how often the example is forgotten through the course of training.

\section{Problem Formulation and Methodology}
\label{sec:problem-formulation-and-methodology}
% \subsection{Notation}

% \label{prelims}
\textbf{Notation. }
Denote the train dataset $\{ (x_i, y_i) \}_{i=1}^{|\mc{D}|}$ as $\mc{D}$  and the validation dataset $\{ (x_i, y_i) \}_{i=1}^{|\mc{V}|}$ as $\mc{V}$. $\model$ denotes a machine learning model parameterized by $\theta \in \mathbb{R}^p$, where $\mathbb{R}^p$ is the parameter space. Let $\ell$ denote an arbitrary loss function. Define the element wise loss function $\ell_i(\theta) := \ell(\model(x_i), y_i)$. Denote the loss on the whole dataset $\mc{D}$ as $\mc{L}_{\mc{D}} (\theta) := \sum_{i \in \mc{D}} \ell_i(\theta)$. We use $\mc{M}(\ldots)$ to denote a differentially private mechanism in the following discussion.

\subsection{Problem Formulation}

We start by specifying our objective function based on GLISTER by \citet{glister},

\begin{equation}
    \label{eqn:bilevel-optimization}
    \underset{\substack{S \subset \mc{D} , |S| \le k}}{\arg \min} \; \mc{L}_{\mc{V}} (\underset{\theta}{\arg \min} \; \mc{L}_{S}(\theta))
\end{equation}

The overall objective consists of two optimization problems. The inner problem optimizes over the model parameters $\theta$, while the outer problem optimizes the val loss over the space of cardinality constrained subsets $S \subseteq \mc{D}$ in order to improve model generalization. It is infeasible to solve the above optimization problem directly for general loss functions and we approximate it in the following way. We iterate over the inner and the outer optimization. The inner optimization yields a model $\theta^*(S)$ for a fixed subset $S$. While the outer problem returns the optimal subset $S^*(\theta)$ given fixed model parameters $\theta$. Solving the inner problem involves gradient descent model training of $\model$ on subset $S$. The outer problem is of combinatorial nature and cannot be solved directly. \citet{glister} prove that monotone submodular proxy exists for optimizing the outer objective for multiple choice of loss functions and use a greedy algorithm \citep{SGA} to quickly extract a training subset.

\begin{figure}[t]
    \centering
    % \subfloat[CIFAR10 $\varepsilon=1$]{
    %     \includegraphics[width=0.4\linewidth]{img/cifar10_main_results_epsilon_1.pdf}
    % } 
    \subfloat[CIFAR10 $\varepsilon=3$]{
        \includegraphics[width=0.23\linewidth]{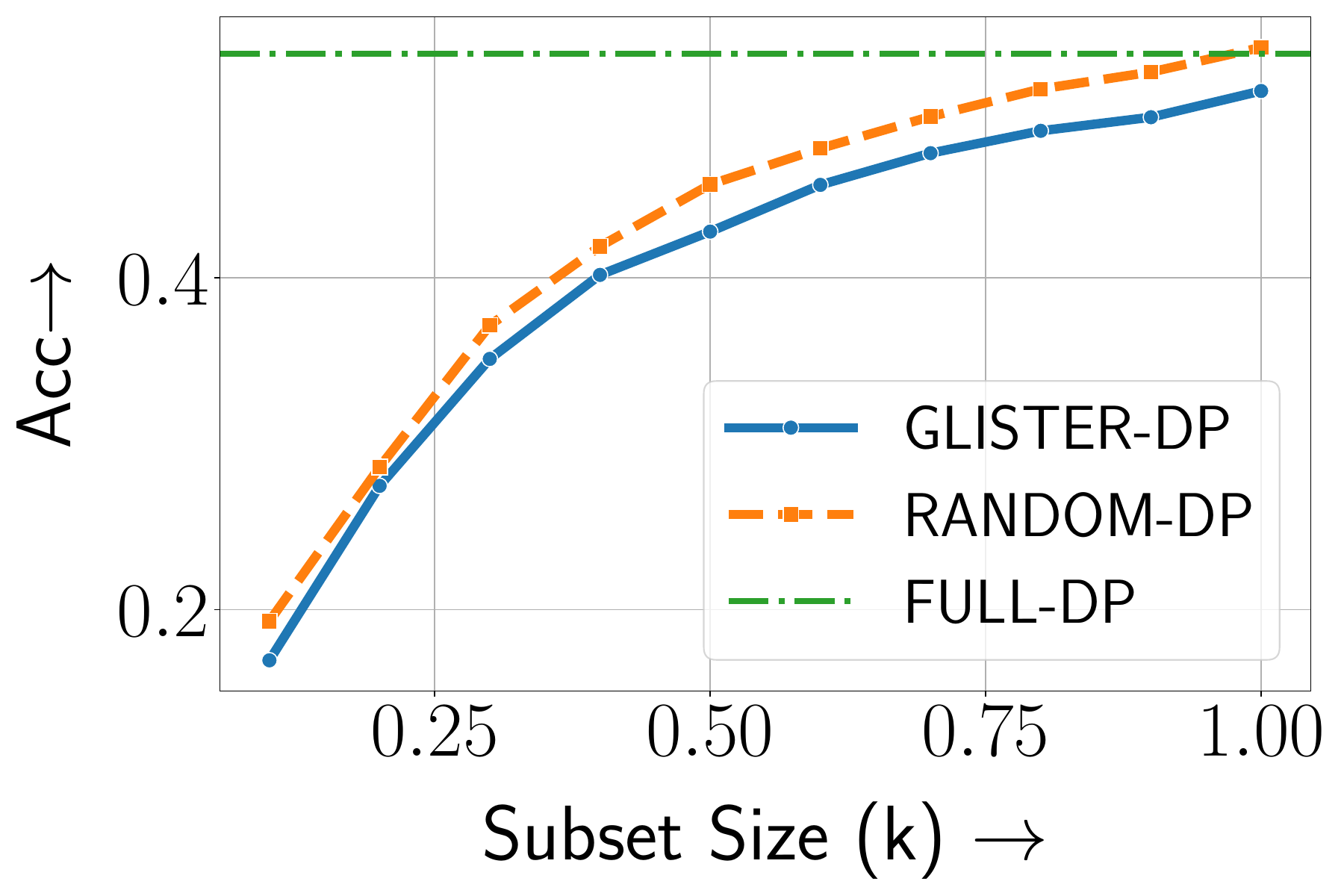}
    } 
    \subfloat[CIFAR10 $\varepsilon=8$]{
        \includegraphics[width=0.23\linewidth]{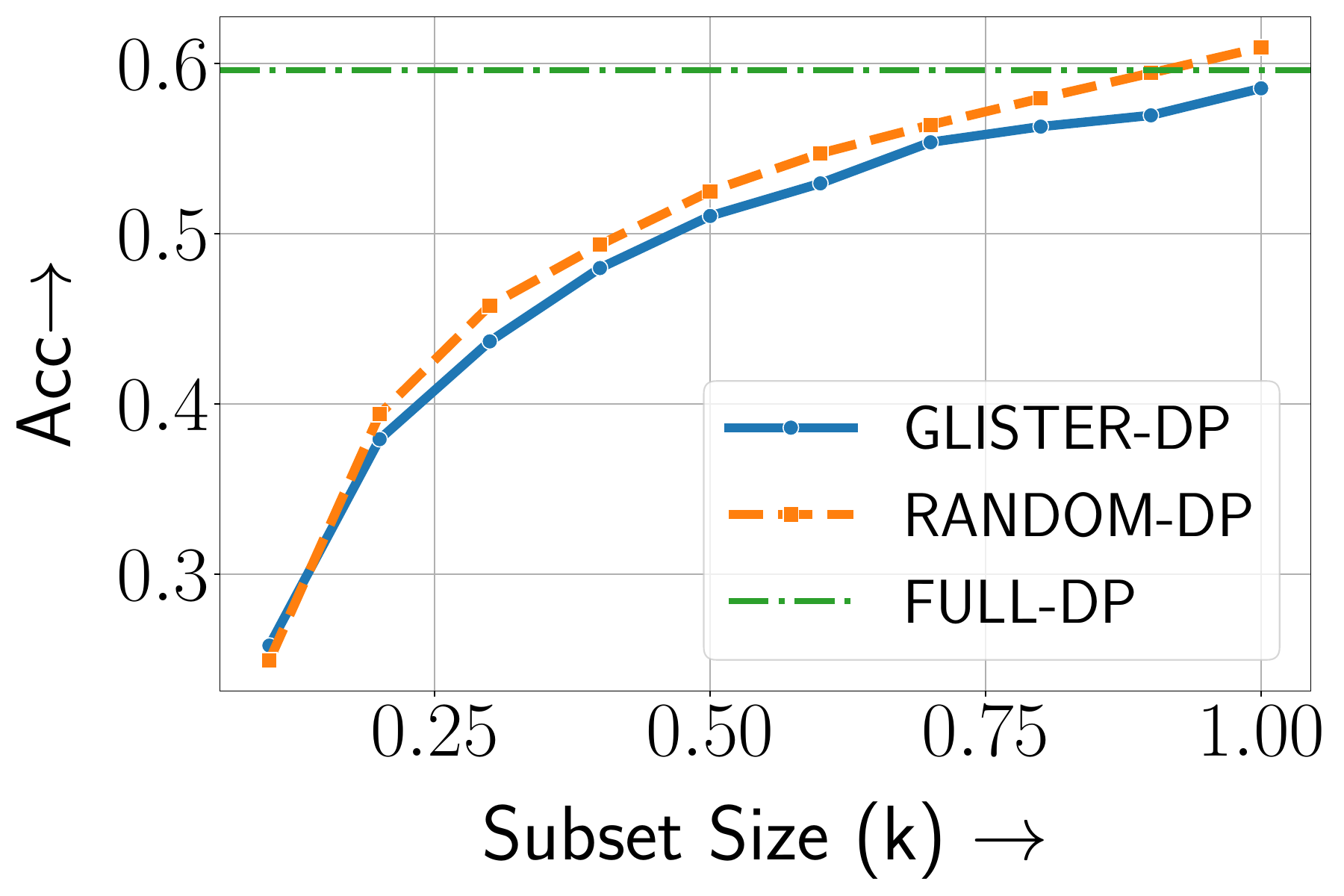}
    }
    % } \\
    % \subfloat[MNIST $\varepsilon=1$]{
    %     \includegraphics[width=0.4\linewidth]{img/MNIST_main_results_epsilon_1.pdf}
    % }
    \subfloat[MNIST $\varepsilon=3$]{
        \includegraphics[width=0.23\linewidth]{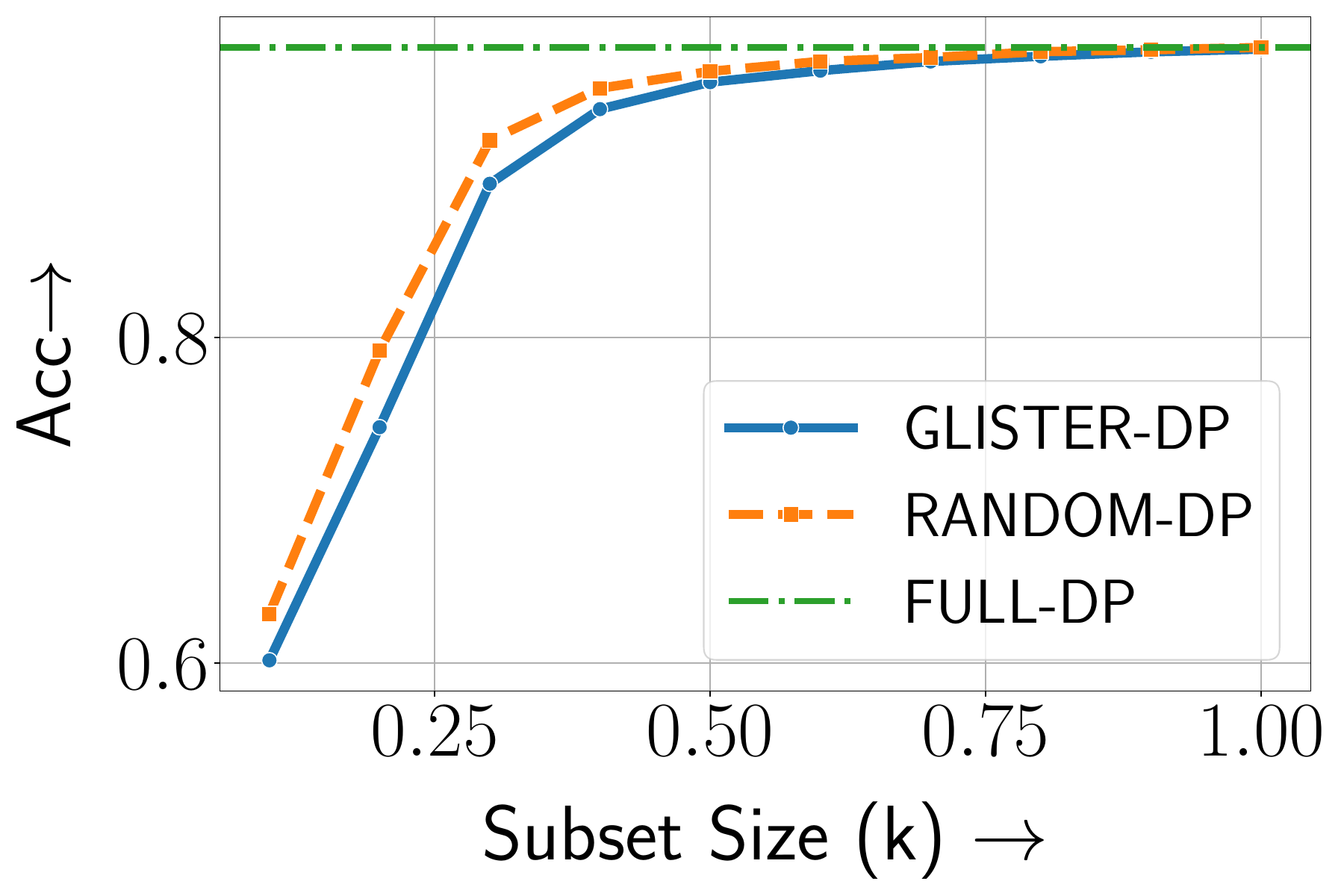}
    }
    \subfloat[MNIST $\varepsilon=8$]{
        \includegraphics[width=0.23\linewidth]{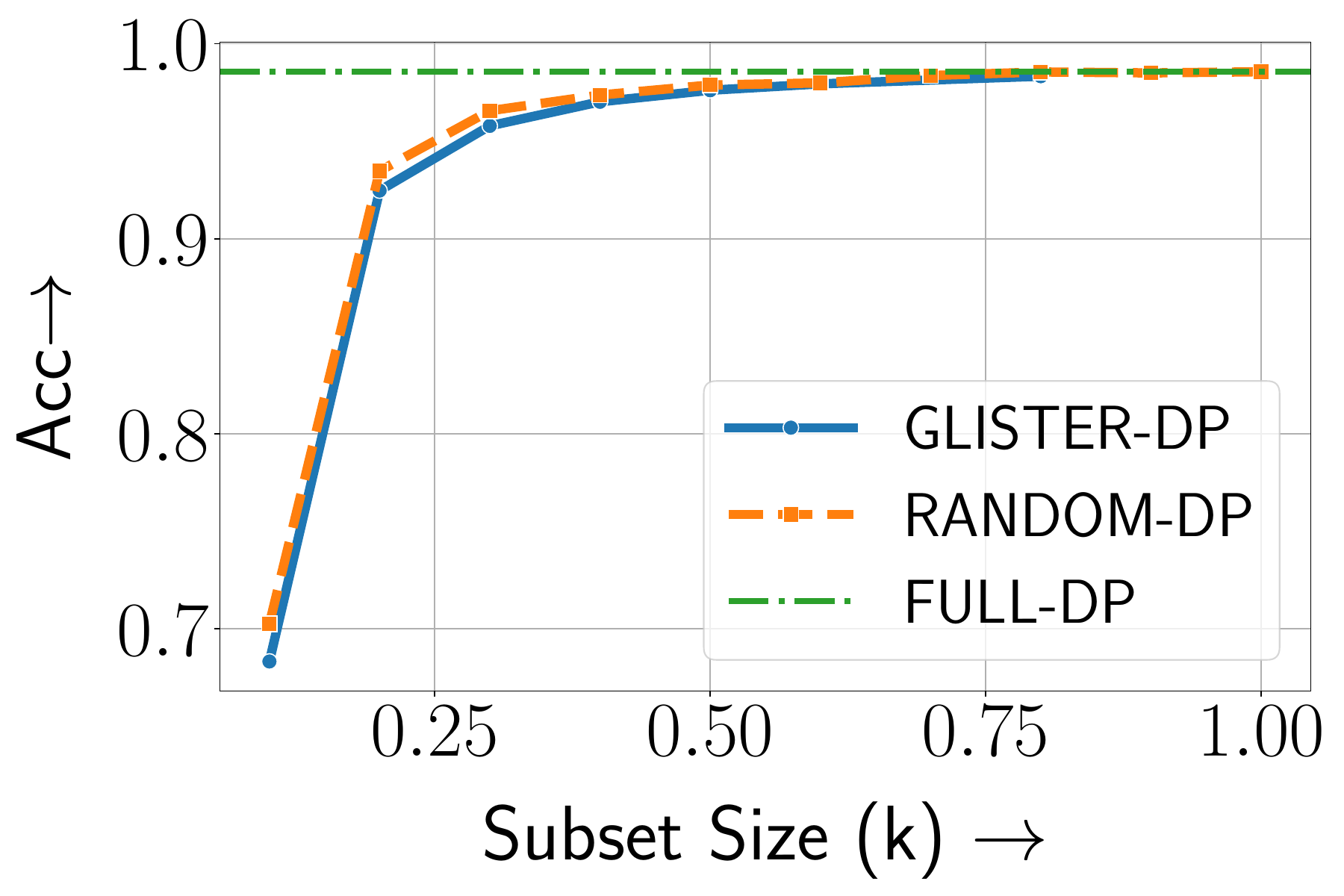}
    }
    \caption{Performance of GLISTER-DP, RANDOM-DP and FULL-DP on test set for CIFAR10 and MNIST with $\varepsilon \in \{3,8\}$ across various choices of subset size $k$ as a fraction of $|\mc{D}|$}
    \label{fig:main-results}
\end{figure}

\subsection{Differentially Private Data Efficient Training}
As discussed perviouslt, the training procedure iterates over model training and subset selection. We describe how we adapt this non private training method to a private version.

\textbf{Differentially Private Training Phase. }
We use the DP-SGD algorithm \citep{dp-sgd} during the training phase. At time step $t$, we use DP-SGD to update model parameters $\theta^t$ by training on the subset $S^t$.
The source of privacy leakage during training is through the gradient computation $g(\theta, S) := \nabla_\theta \mc{L}_S(\theta)$. DP-SGD performs \textbf{gradient clipping} and adds \textbf{multidimensional Gaussian noise} to the gradients. The noise scale $\sigma_g$ is based on the privacy parameters $\varepsilon$ and $\delta$ and also depends on the maximum $l_2$ norm of gradients which is bounded to some constant $C$. The privacy accountant tracks the degradation of privacy throughout the training phase. We denote the DP training mechanism $\mc{M}_g(\theta, S, g(\cdot)) := g(\theta, S) + p$ where $p \sim \mc{N}(0, \sigma_g)$.

\textbf{Differentially Private Subset Selection Phase. }
The subset selection procedure is reformulated as a submodular maximization problem by \citet{glister} which can be solved using the stochastic greedy algorithm due to \citet{SGA}. At its core, an optimal subset $S$ that approximately ($(1-1/e)$ approximation guarantee) maximizes a submodular objective function $F$ can be found by greedily choosing an element $e$ with maximum gain $F(S \cup{e}) - F(S)$ in a sequential manner. We outline the detailed algorithm for differentially private subset selection in Appendix~\ref{appn:algo} based on the DP submodular maximization algorithm 
 by \citet{dp-submod}, using the exponential mechanism \citep{exp-mech} for differential privacy as its core primitive. The argmax step in the greedy algorithm gets replaced by a sampling step based on the exponential mechanism. Overall, the optimization procedure is a k-fold composition of exponential mechanisms, yielding one element at each step. \citet{dp-submod} provide privacy bounds along with approximation guarantees for the overall differentially private submodular optimization algorithm. Denote the DP subset selection mechanism $\mc{M}_{ss}(\theta, \mc{D}, F(\cdot))$, composed of multiple exponential mechanisms.

\textbf{Algorithm. }
The detailed description of our training algorithm can be found in Appendix~\ref{appn:algo}. We adapt GLISTER \citep{glister} by replacing training with DP-SGD and subset selection with the DP submodular maximization algorithm by \citet{dp-submod}. We use basic composition for privacy accounting of the two heterogeneous mechanisms $\mc{M}_{g}$ for training and $\mc{M}_{ss}$ for subset selection. We refer to our method as GLISTER-DP in our experiments.

\textbf{Privacy Accounting. }
Privacy accounting during training phase is due to the numerical composition algorithm by \citet{prv-accountant}, and runs in $O(\sqrt{k})$ time for k-fold adaptive composition of homogeneous DP mechanisms. The privacy accounting during the subset selection phase is based on the analysis given by \citet{dp-submod}.  We split the total privacy budget into two parts, $\varepsilon_g$ for training and $\varepsilon_{\text{ss}}$ for data subset selection. This follows from the basic composition theorem of DP mechanisms.

% For an overall budget of $\varepsilon_{ss}$ in the subset selection phase, the greedy algorithm for DP submodular maximization sequentially selects $k$ elements through $k$ sequential applications of the $(\varepsilon_0, \delta)$-DP exponential mechanisms $\mc{O}^{em}$, where $\varepsilon_0$, $\varepsilon_{ss}$ and $k$ are related as $\varepsilon_{ss} = k\varepsilon_0^2/2 + \varepsilon_0 \cdot \sqrt{2kln(1/\delta)}$ for any choice of $\delta \gt 0$.

\section{Experiments and Discussion}

\begin{wrapfigure}{r}{0.35\textwidth}
    \vspace{-0.5cm}
    \begin{center}
        \includegraphics[width=0.33\textwidth]{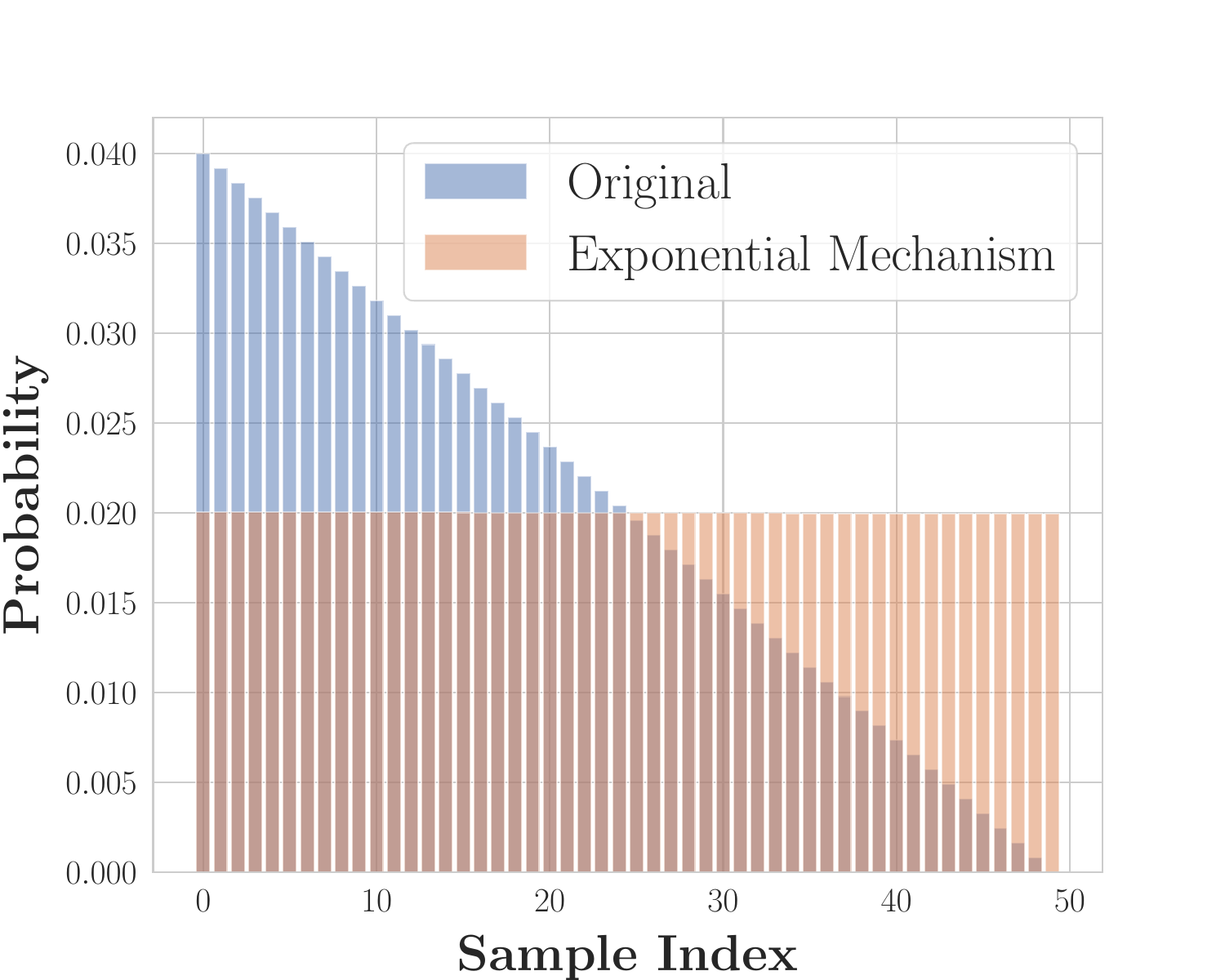}
    \end{center}
    \caption{Comparison of original distribution with the one that exponential mechanism samples from. The plot is generated by resampling the true gains and noisy gains and normalizing to produce a valid probability distribution.}
    \label{fig:distr-noise}
    \vspace{-0.3cm}
\end{wrapfigure}

\textbf{Datasets and Baselines. }
We experiment with two real world image datasets CIFAR10 and MNIST. We also provide results on class imbalanced synthetic datasets in Appendix~\ref{appn:imbalanced-synthetic}. We compare our GLISTER-DP approach with two baselines. (1) RANDOM-DP selects a training subset $S \subseteq \mc{D}$ of size $k$ uniformly at random. RANDOM-DP does not incur any privacy cost during subset selection phase, and the whole budget goes to private training. (2) FULL-DP always trains on the full dataset, and provides a reference for comparison. We test the performance of our methods across various values of $k$, choosing $k \in [0,1]$ as a fraction of $\mc{D}$ and for $\varepsilon \in \{3, 8\}$. Our experiments can be reproduced by running our \href{https://anonymous.4open.science/r/DP-SubSel-70C3/}{code}.

\textbf{Main Results. }
We discuss the main results of our experiments shown in Figure~\ref{fig:main-results}. We observe that full training beats both subset selection methods. We also observe that RANDOM-DP outperforms GLISTER-DP for all values of $\varepsilon$ and for both dataset MNIST and CIFAR10. As discussed in Section \ref{sec:problem-formulation-and-methodology}, GLISTER-DP splits the total privacy budget $\varepsilon_{\text{total}}$ into two parts, allocating $\varepsilon_g$ for training and $\varepsilon_{ss}$ for subset selection. Correspondingly, the training noise scale for GLISTER-DP is significantly higher than RANDOM-DP.

\begin{wrapfigure}{r}{0.35\textwidth}
    \vspace{-0.3cm}
    \begin{center}
        \includegraphics[width=0.33\textwidth]{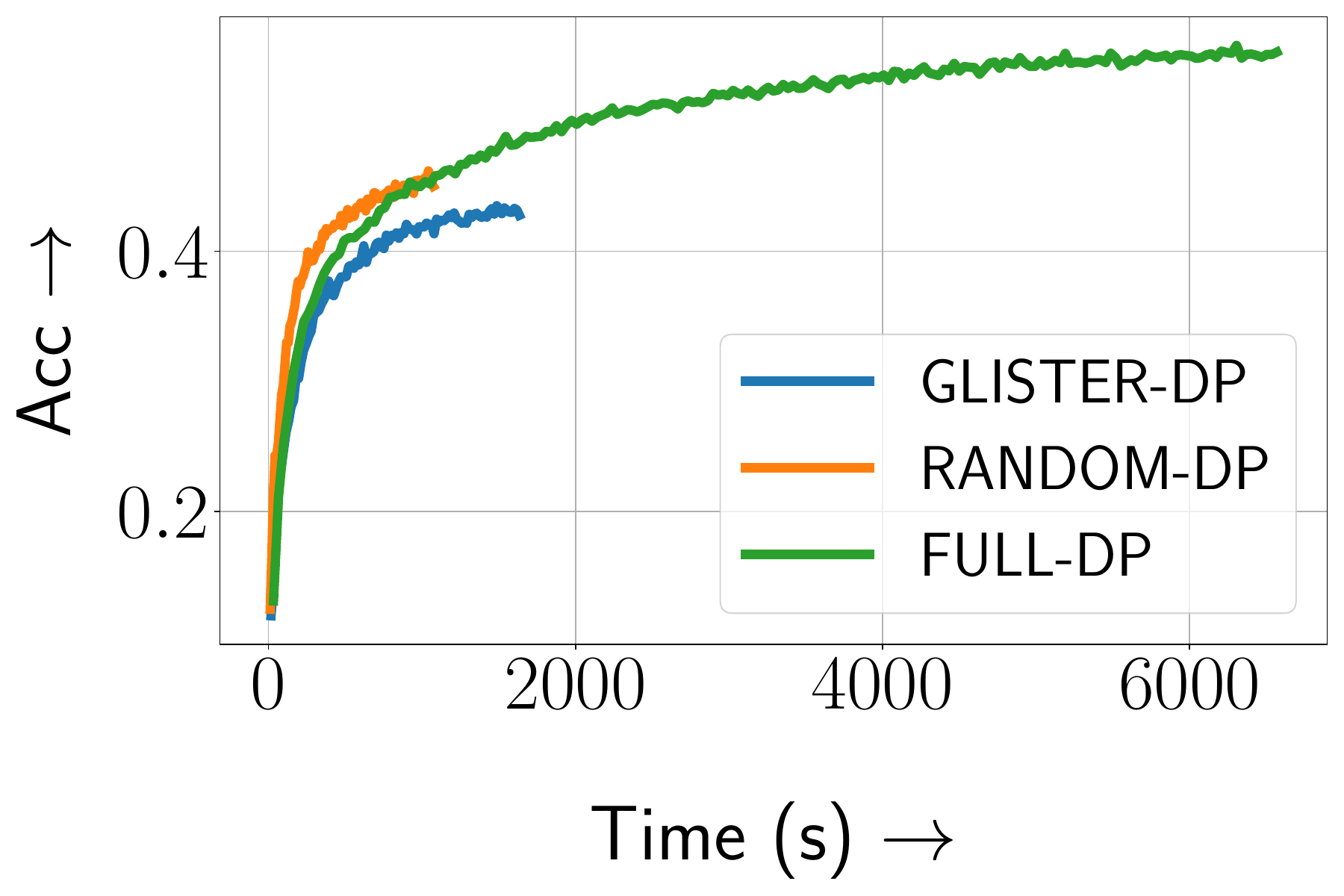}
    \end{center}
    \caption{Training convergence for each method on CIFAR10, $\varepsilon=3$ and $k=0.5|\mc{D}|$}.
    \label{fig:timing-plot}
    \vspace{-0.3cm}
\end{wrapfigure}

During the subset selection phase, GLISTER-DP must make up for the disadvantage of noisier training by choosing a high quality training subset. We show that this is not the case, with help of Figure~\ref{fig:distr-noise}. In the figure, we provide a comparison between the true distribution of the gains of each element and the distribution that the exponential mechanism samples from. The sampling distribution is extremely noisy and it is almost equivalent to sampling elements uniformly at random. Empirically, we observe that the privacy budget $\varepsilon_{ss}$ is too restrictive to yield a good training subset and the generated subset is near random. This explains the loss in performance of GLISTER-DP.

\textbf{Timing Analysis. }
In Figure~\ref{fig:timing-plot}, we show the training convergence of each method for $k = 0.5 |\mc{D}|$ (with $k=|\mc{D}|$ for FULL-DP). We observe that RANDOM-DP converges quicker than FULL-DP and GLISTER-DP is the slowest to converge. We observe this trend across all values of $k$ and show this in Appendix~\ref{appn:timing-analysis}.   
% GLISTER chooses an element during the subset selection phase if the direction of the gradient matches with the direction of the val gradient. In every iteration, the element that maximizes the gain of the submodular gains function is chosen. This argmax operation gets replaced by the exponential mechanism for our private version which samples from the \textit{noised distribution of gains} rather than choosing the top element. In figure \ref{fig:distr-noise} we show the original distribution of gains and the distribution after adding noise. From our experiments, we clearly see that the noise during subset selection is too high and the chosen subset is as good as random. This explains the degradation in performance of GLISTER throughout.

\textbf{Other Experiments. }
In the appendix, we discuss experiments with imbalanced datasets Appendix~\ref{appn:imbalanced-synthetic}. We induce imbalance in real world datasets as well as generate synthetic datasets. We observe that our approach GLISTER-DP performs better than baselines. We also discuss the change in training performance by varying budget allocation between training and subset selection.

\section{Conclusion}

In this work, we investigate the potential interaction between data efficient deep learning with differential privacy. To this end, we develop GLISTER-DP, a method for data efficient model training in the private setting based on GLISTER \citep{glister}. We use DP-SGD \citep{dp-sgd} for training and differentially private submodular maximization algorithm by \citet{dp-submod} for subset selection. The most essential part of data efficient model training is efficient search of good quality data for training. We empirically observe, that differential privacy poses a significant challenge on the data subset search problem as the privacy budget is too restrictive, rendering it impractical.

% \begin{minipage}{0.4\textwidth}    
%             \includegraphics[width=0.33\textwidth]{img/timing_plots_train_frac_0_5_eps_3.pdf} \\
%             \includegraphics[width=0.33\textwidth]{img/Accuracy_Vs_Alloc.pdf} \\
%             \includegraphics[width=0.33\textwidth]{img/noised_distributions.pdf}
%         \label{fig:enter-label}
% \end{minipage}

% \begin{minipage}{0.4\textwidth}    
%     \begin{figure}[b]
%         \centering
%         \subfloat[timing plot]{
%             \includegraphics[width=0.33\textwidth]{img/timing_plots_train_frac_0_5_eps_3.pdf}
%         }
%         \subfloat[Allocation Rate]{
%             \includegraphics[width=0.33\textwidth]{img/Accuracy_Vs_Alloc.pdf}
%         }
%         \subfloat[\small Original distribution of gains compared to the noised distribution. Actual distribution is over 45000 elements in the train set, that has been resampled here for illustration.]{
%             \includegraphics[width=0.33\textwidth]{img/noised_distributions.pdf}
%         }
%         \caption{Caption}
%         \label{fig:enter-label}
%     \end{figure}
% \end{minipage}



\bibliography{iclr2025_conference}
\bibliographystyle{iclr2025_conference}

\newpage
\appendix
\section*{Appendix}

\section{Experimental Details}
\label{appn:exp-details}
The timing numbers are reported on the runs on NVIDIA A6000 GPUs. We do not perform hyperparameter tuning, and run all methods on the same set of hyperparameters in order to reduce computation and expenditure of privacy budget for the same. Throughout our experiments, hyperparameters are chosen so that the noise scale $\sigma$ remains significantly above the "privacy wall" \citep{dp-tan-burn} and yet allows for model training.

\section{GLISTER vs GLISTER-DP}
\label{appn:algo}

In the following, we compare the original GLISTER algorithm with the DP variant GLISTER-DP. The notable changes in the algorithm are inclusion of the privacy accountants $\PAtrain$ and $\PAss$ and replacement of the normal training with DP-SGD based private training with $\varepsilon_g$ budget and greedy submodular maximization with the DP version for $\varepsilon_{\text{ss}}$ budget.

\begin{minipage}{0.44\textwidth}
    
\begin{algorithm}[H]
    \caption{GLISTER}
    \begin{algorithmic}
    \vspace{0.4cm}
        \State \textbf{Input: } Trainset: $\mc{D}$, valset: $\mc{V}$, initial subset: $S^0$, initial model: $\theta^0$. LR: $\eta$, epochs: $T$, batch size $B$, selection interval: $L$.
        \State \textbf{Output: } Final model $\theta^T$, Final subset $S^T$.
        \State
        \For{epoch in $1 \ldots T$}
            \If{epoch \% L == 0}
                \State $S^{t+1} \gets$ GreedyAlgo($\mc{D}, \mc{V}, \theta^t, \eta$)
            \Else
                \State $S^{t+1} \gets S^t$
            \EndIf
            \State $\theta^{t+1} \gets$ Train($\theta^t, S^{t+1}$)
        \EndFor
        \State \Return $\theta^T, S^T$
    \vspace{0.4cm}
    \end{algorithmic}
\end{algorithm}
\end{minipage}
\hfill
\begin{minipage}{0.54\textwidth}
    
\begin{algorithm}[H]
    \caption{GLISTER-DP}
    \begin{algorithmic}
        \State \textbf{Input: } Trainset: $\mc{D}$, valset: $\mc{V}$, initial subset: $S^0$, initial model: $\theta^0$. LR: $\eta$, epochs: $T$, batch size $B$, selection interval: $L$, privacy budget $(\varepsilon, \delta)$, allocation ratio $r$
        \State \textbf{Output: } Final model $\theta^T$, Final subset $S^T$.
        \State $\varepsilon_{train} \gets \varepsilon\cdot r$
        \State $\varepsilon_{ss} \gets \varepsilon\cdot (1-r)$
        \State Initialize $\PAtrain \gets \textbf{Accountant}(T, B, \varepsilon_{g})$
        \State Initialize $\PAss \gets \textbf{Accountant}(T, L, \varepsilon_{ss})$
        \For{epoch in $1 \ldots T$}
            \If{epoch \% L == 0}
                \State $S^{t+1} \gets$ \textbf{DP-GreedyAlgo}($\mc{D}, \mc{V}, \theta^t, \eta, \PAss$)
            \Else
                \State $S^{t+1} \gets S^t$
            \EndIf
            \State $\theta^{t+1} \gets$ \textbf{DP-Train}($\theta^t, S^{t+1}, \PAtrain$)
        \EndFor
        \State \Return $\theta^T, S^T$
        
    \end{algorithmic}
\end{algorithm}

\end{minipage}

Due to restricted privacy budget, we perform subset selection every $L$ epochs and use the subset for training for the next $L$ epochs.

% \begin{algorithm}
% \caption{DP-GreedyAlgo($U, V, \theta^0, \eta, k, r, \lambda, R, \epsilon$)}
% \begin{algorithmic}[1]
%     \State Initialize $S = \emptyset$, $U = \mathcal{U}$, $t = 0$
%     \While{$t < r$}
%         \For{all $e \in U$}
%             \State Set $\theta_e^{(t)} = \theta^{(t)} + \eta \nabla_\theta LL_T(e, \theta) |_{\theta^{(t)}}$
%             \State Set $\tilde{G}_{\theta}(e) = G_{\theta_e^{(t)}}(e | S^k) + \lambda R(e | S^k)$
%         \EndFor
%         \State Set $X = \emptyset$
        
%         \For{$i = 1$ to $k/r$}
%             \State Sample $e^*$ from $U$ with probability:
%             $\Pr(e^*) \propto \exp\left(\frac{\epsilon}{2\Delta} \tilde{G}_\theta(e^*)\right)$
%             \State $X = X \cup \{e^*\}$
%             \State $U = U \setminus \{e^*\}$
%         \EndFor
%         \State $S_t = S_t \cup \{X\}$
%         \State $S = S \cup S_t$
%         \State Update $\theta^{(t+1)} = \theta^{(t)} + \sum_{e \in S_t} \nabla_\theta LL_T(e, \theta) |_{\theta^{(t)}}$
%         \State $t = t + 1$
%     \EndWhile
%     \State \Return $S$
% \end{algorithmic}
% \end{algorithm}

\newpage
\section{Class Imbalanced Synthetic Datasets}
\label{appn:imbalanced-synthetic}

% \begin{wrapfigure}{R}{0.4\textwidth}
%     \begin{center}
%         \includegraphics[width=0.38\textwidth]{image.png}
%     \end{center}
%     \caption{Caption}
% \end{wrapfigure}
\textbf{Real world datasets with induced class imbalance. }
We first present results on class imbalanced real world datasets. The number of samples for a class vary between 80 percent to 100 percent and is created artificially on the datasets MNIST, CIFAR10 and CIFAR100. We show the results in Table~\ref{tab:imbalance}. We see that GLISTER-DP outperforms RANDOM-DP on these imbalanced datasets and underlines the utility of the subset selection methods under class imbalanced settings.

\begin{table}[h]
\begin{tabular}{|l|l||c|c|c|c|c|l|}
\hline
\textbf{Dataset}           & \textbf{Method} & \textbf{$\epsilon$} & \textbf{$k= 0.1|\mc{D}|$} & \textbf{$k= 0.2|\mc{D}|$} & \textbf{$k= 0.3|\mc{D}|$} & \textbf{$k= 0.4|\mc{D}|$} & \textbf{$k= 0.5|\mc{D}|$} \\ \hline
\multirow{2}{*}{MNIST}     & RANDOM-DP          & 3.0        & 0.6982                & 0.9103                & 0.9474                & 0.9602                & 0.9649                \\ \cline{2-8} 
                           & GLISTER-DP         & 3.0        & 0.7231                & 0.9155                & 0.9490                & 0.9609                & 0.9657                \\ \cline{2-8} 
                           & RANDOM-DP          & 8.0        & 0.8979                & 0.9599                & 0.9707                & 0.9739                & 0.9760                \\ \cline{2-8} 
                           & GLISTER-DP         & 8.0        & 0.9059                & 0.9617                & 0.9710                & 0.9744                & 0.9768                \\ \hline
\multirow{2}{*}{CIFAR-100} & RANDOM-DP          & 3.0        & 0.0162                & 0.0249                & 0.0510                & 0.0664                & 0.0854                \\ \cline{2-8} 
                           & GLISTER-DP         & 3.0        & 0.0162                & 0.0274                & 0.0483                & 0.0734                & 0.0829                \\ \cline{2-8} 
                           & RANDOM-DP          & 8.0        & 0.0344                & 0.0838                & 0.1053                & 0.1337                & 0.1385                \\ \cline{2-8} 
                           & GLISTER-DP         & 8.0        & 0.0362                & 0.0810                & 0.1118                & 0.1234                & 0.1433                \\ \hline
\multirow{2}{*}{CIFAR-10}  & RANDOM-DP          & 3.0        & 0.2872                & 0.3631                & 0.4174                & 0.4460                & 0.4598                \\ \cline{2-8} 
                           & GLISTER-DP         & 3.0        & 0.2751                & 0.3719                & 0.4189                & 0.4509                & 0.4669                \\ \cline{2-8} 
                           & RANDOM-DP          & 8.0        & 0.3894                & 0.4523                & 0.4856                & 0.5026                & 0.5317                \\ \cline{2-8} 
                           & GLISTER-DP         & 8.0        & 0.3878                & 0.4564                & 0.4808                & 0.5111                & 0.5494                \\ \hline
\end{tabular}
\caption{Comparison of performance of RANDOM-DP and GLISTER-DP on mild class imbalance datasets across fraction of training budget}
\label{tab:imbalance}
\end{table}

\begin{figure}[h]
    \centering
    \includegraphics[width=0.6\linewidth]{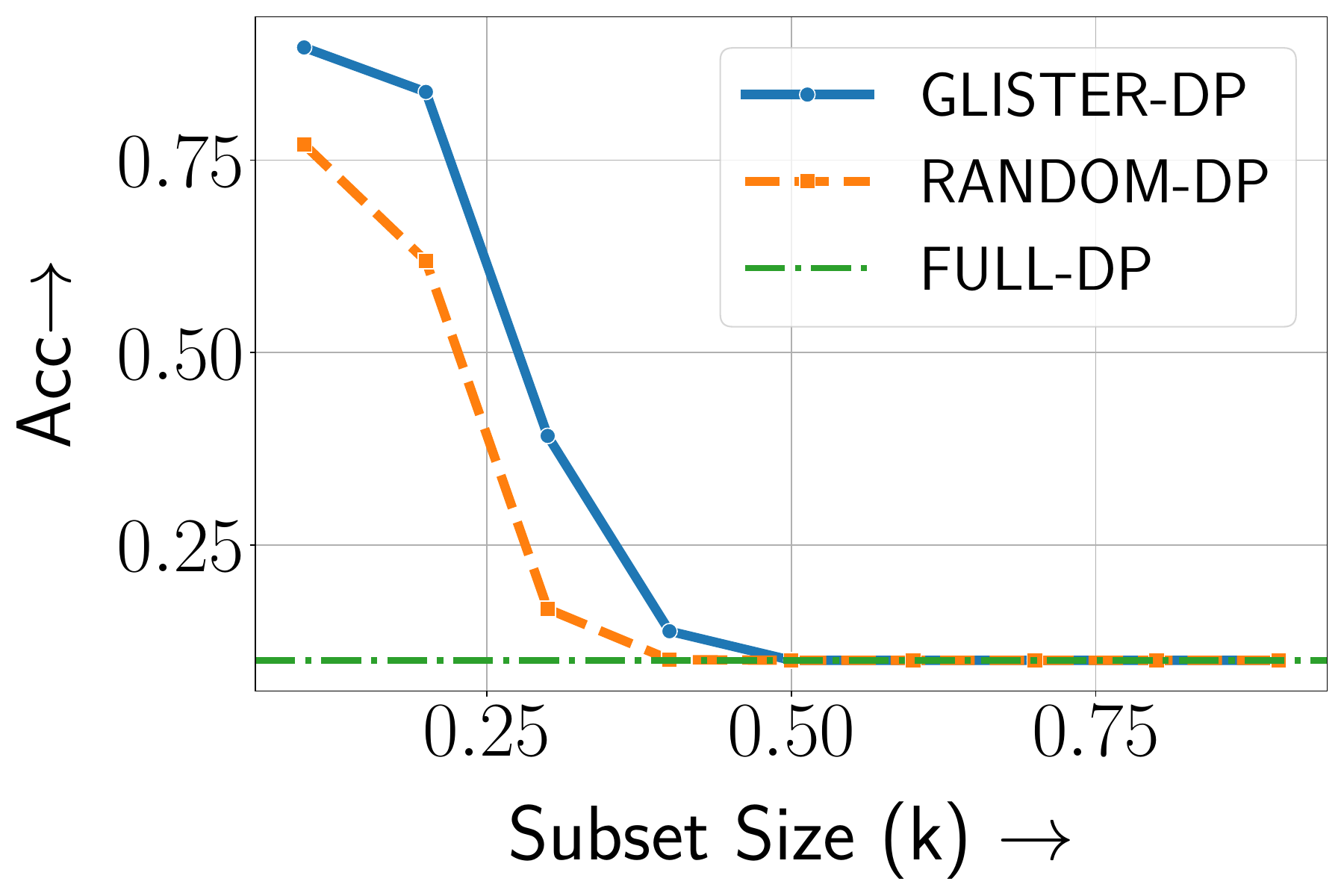}
    \caption{Performance comparison on highly imbalanced synthetic dataset. }
    \label{fig:imbalanced-synthetic}
\end{figure}

\textbf{Experiments with highly imbalanced synthetic dataset. }
Next we provide results on an imbalanced synthetic dataset to illustrate the applicability of data subset selection methods. We create a synthetic dataset such that it has significant train, val and test distribution shift. The synthetic dataset contains $N=5000$ examples, each example having $m=10$ features and the dataset contains $2$ classes. Train dataset has a class imbalance ratio of 1:9, val dataset imbalance ratio is 6:4 and test dataset has an imbalance ratio 9:1. Under these settings, GLISTER-DP has a significant edge over other baselines since the choice of training subset for GLISTER-DP is informed based on the val set as can be seen in Figure~\ref{fig:imbalanced-synthetic}. As the size of the train subset increases, the performance of both GLISTER-DP and RANDOM-DP become equivalent to FULL-DP.

\newpage
\section{Timing Analysis}
\begin{figure}[h]
    \centering
    \includegraphics[width=0.6\textwidth]{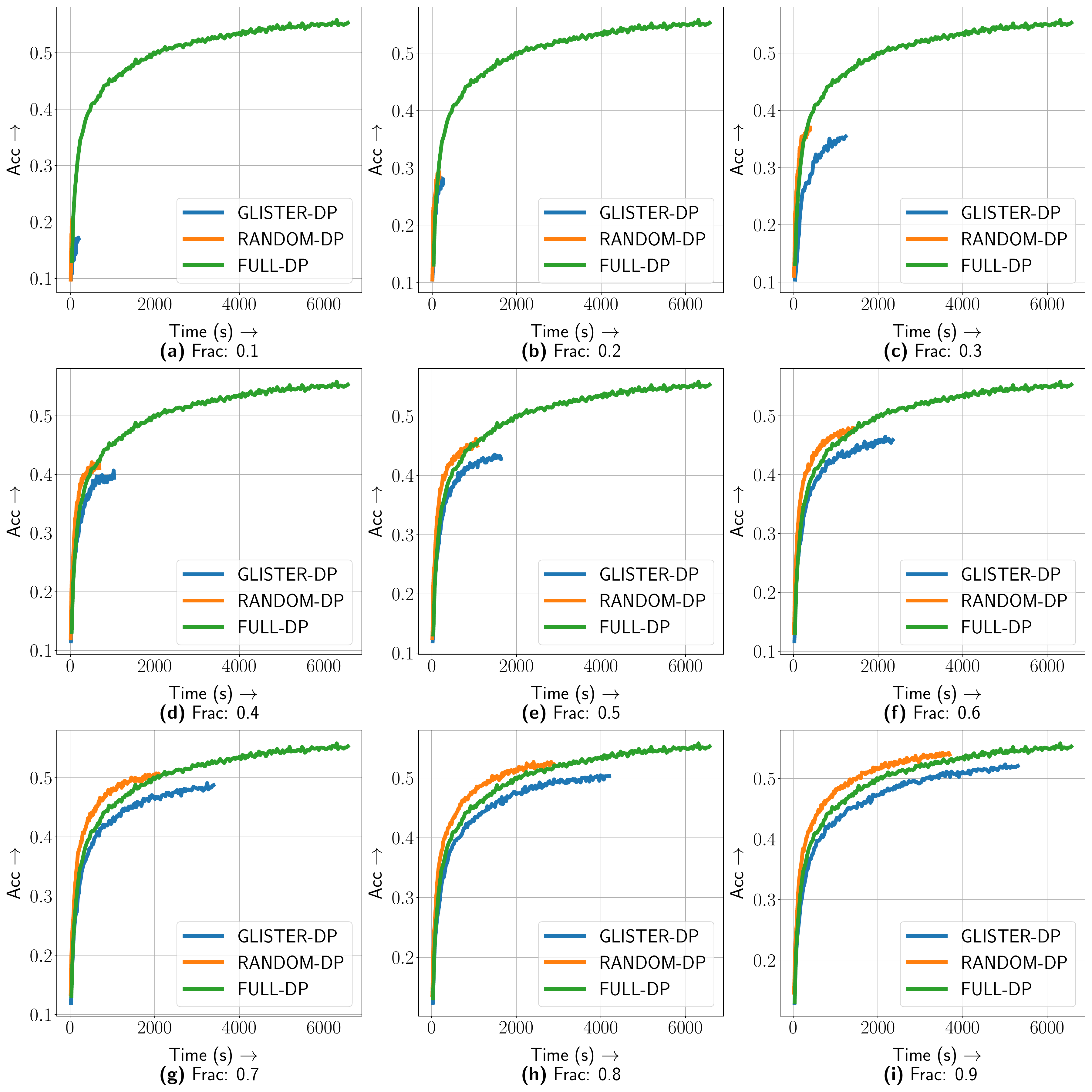}
    \caption{Training convergence plot of GLISTER-DP, FULL-DP and RANDOM-DP across different fractions of training budget on CIFAR-10 $\epsilon=3$ }
    \label{fig:timing-plot-all}
\end{figure}

We provide the timing analysis of convergence of all the three methods in Figure~\ref{fig:timing-plot-all}. The following experiment is conducted for CIFAR10 with privacy budget $\varepsilon = 3$ and $\delta=10^{-5}$. We observe that the training on the random subset give fastest convergence in general. GLISTER-DP converges the slowest across all choices of $k$.

\label{appn:timing-analysis}

% \begin{wrapfigure}{R}{0.9\textwidth}
%     \begin{center}
%         \includegraphics[width=0.8\textwidth]{img/timing_plots_train_frac_0_1_to_0_9_eps_3.pdf}
%     \end{center}
%     \caption{Caption}
% \end{wrapfigure}

\section{Experiments with Allocation Rate. }
\label{appn:alloc}

\begin{figure}[h]
    \centering
    \includegraphics[width=0.5\linewidth]{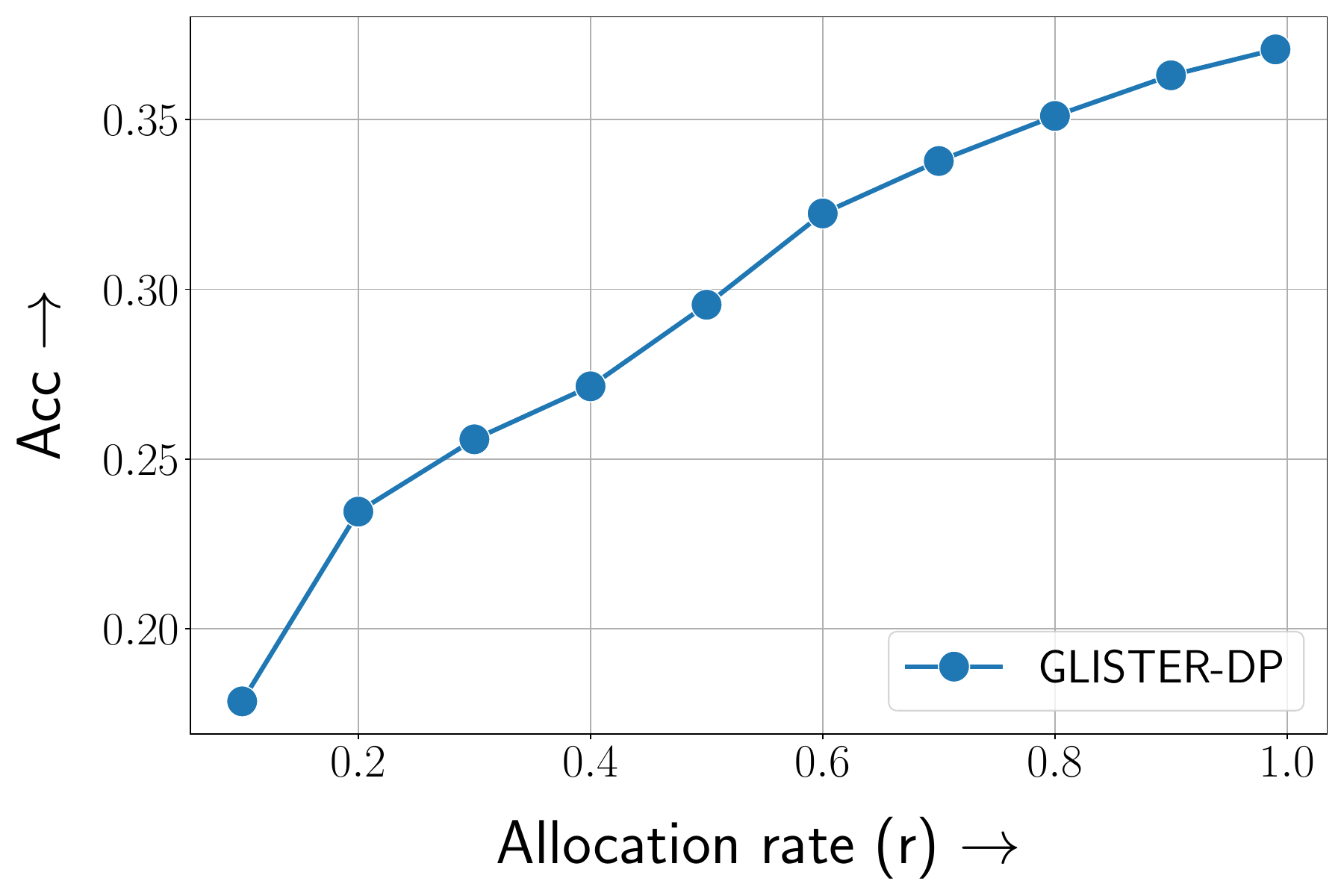}
    \caption{Performance of GLISTER-DP across various choices of budget allocation. $r=0.1$ corresponds to $\varepsilon_g=0.1 \cdot\varepsilon_{\text{total}}$}
    \label{fig:alloc}
\end{figure}

In Figure~\ref{fig:alloc} we show the effect of budget allocation for GLISTER-DP. Lower allocation rate corresponds to the $\varepsilon_g$ being low, reducing the training privacy budget. We see that the performance of GLISTER-DP monotonically increases as we increase the training budget. Allocating higher budget for subset selection does not improve the subset quality to mitigate the performance degradation during model training. We observe that there is no \textit{sweet spot} in the trade-off between $\varepsilon_g$ and $\varepsilon_{ss}$ and that it is always better to spend privacy budget on training rather than choosing a subset.

\end{document}